# Deep Speech Based End-to-End Automated Speech Recognition (ASR) for Indian-English Accents

Priyank Dubey
Indian Institute of Science Education and Research (IISER)
Berhampur, Odisha
priyank19@iiserbpr.ac.in

Bilal Shah
Centre for Artificial Intelligence and Robotics (CAIR)
Defence Research and Development Organization (DRDO)
Bengaluru, Karnataka
bilalshah@gmail.com

## Abstract

Automated Speech Recognition (ASR) is an interdisciplinary application of computer science and linguistics that enable us to derive the transcription from the uttered speech waveform. It finds several applications in Military like High-performance fighter aircraft, helicopters, air-traffic controller. Other than military speech recognition is used in healthcare, persons with disabilities and many more. ASR has been an active research area. Several models and algorithms for speech to text (STT) have been proposed. One of the most recent is Mozilla Deep Speech, it is based on the Deep Speech research paper by Baidu. Deep Speech is a state-of-art speech recognition system is developed using end-to-end deep learning, it is trained using well-optimized Recurrent Neural Network (RNN) training system utilizing multiple Graphical Processing Units (GPUs). This training is mostly done using American-English accent datasets, which results in poor generalizability to other English accents. India is a land of vast diversity. This can even be seen in the speech, there are several English accents which vary from state to state. In this work, we have used transfer learning approach using most recent Deep Speech model i.e., deepspeech-0.9.3 to develop an end-to-end speech recognition system for Indian-English accents. This work utilizes fine-tuning and data argumentation to further optimize and improve the Deep Speech ASR system. Indic TTS data of Indian-English accents is used for transfer learning and fine-tuning the pre-trained Deep Speech model. A general comparison is made among the untrained model, our trained model and other available speech recognition services for Indian-English Accents.

1. ## Introduction

    In the last few decades, the world has witnessed a great advancement in the technology. Data and Information processing machines have become ubiquitous. The future success is totally relied on the people and machines working together in a collaborative manner. Through such collaborative work, people and machines can enhance each other's complementary skills and strength. In order to completely exploit this collaboration, there must be a proper means of human machine communication. This communication must be proper and flexible to humans and computers. However, the current prevalent mode of inputting the data and information into the machines is either through a keyboard or a mouse. Speech is the primary and flexible mode of communication for humans.

    Speech is a useful expression and has a particular meaning and it is composed of several words and symbols and contain several letters accompanied by voices. Speech has lots of information and full of features, which can be useful for speech recognition. The first work in this field of speech recognition is done by three Bell Laboratories Researchers in 1952. They designed a system called Audrey for single speaker digit recognition. Several other experiments and algorithms have been proposed for ASR system by various companies and researchers. Today, we have commonly available state-of-the-art Automatic Speech Recognition Engines like Apple's Siri, Amazon's Alexa and Google Assistant. However, there are several challenges associated with commonly available ASR like variability of pitch, variability of accents, noises, word speeds.

In 2017 Deep Speech model, a state-of-the-art speech recognition system developed using end-to-end deep learning was presented. It is based on Baidu Research Lab's research paper. Deep Speech tend to perform in challenging noisy environments better than widely used, state-of-the-art commercial speech systems. As it is based on deep learning, the training datasets used to train are mostly comprised of American-English accents. Hence, it performs poorly for all other English accents.

India has always been known as a country with vast diversity. India houses over 19,500 languages or dialects being spoken as a mother tongue (Chandramouli and General, 2011). Even if only official languages are considered, the Eighth Schedule of the Indian Constitution lists 22 official languages. Each of them being spoken by over a million people. Transfer learning approach using most recent Deep Speech model i.e., deepspeech-0.9.3 to develop an end-to-end speech recognition system for Indian-English accents. This paper will present the complete process of fine tuning the Deep Speech system for Indian-English accents. Indic TTS database of Indian-English accents is used for training, which is the largest publicly available resource available for training text-to-speech systems. It is a special corpus of Indian languages covering 13 major languages of India, comprises of 10000+ spoken sentences of English recorded by both Male and Female native speakers from different regions of India.

## 2 Automated Speech Recognition

Automatic Speech Recognition or ASR is a multipurpose technological method that allows humans to input data and information into the computer by simply speaking to the computer interface. It converts continuous audio speech to equivalent text, which is further interpreted by the computer to extract required commands thus creating a kind of natural communication between man and machine. ASR is a modern day need of technology for flexible interaction of humans and machines. There are several challenges associated with the performance of an ASR system such as variability of volume, variability of words speed, variability of speaker, variability of pitch, word continuations and boundaries, and background noises.

For the purpose of speech recognition, the audio speech is treated as speech signal. The ASR system should receive the input speech signal and extract the relevant features for recognition, as features helps in discriminating similar sounds. The speech signal waveform is categorized into segments called frames; from these frames a multi-dimensional feature vector is extracted. The frames act as an analysis ground for speech recognition.

## 3 Applications of ASR

There are several applications of ASR system in various sectors. A number of industries utilize different applications of speech technology today for performing a variety of tasks. ASR system's application in different fields is described below-

**Health Care:** ASR can be implemented for the medical documentation, doctors can utilize to make treatment notes, to capture and log patient diagnoses. The application can be further extended to medical machinery and surgical assistance.

**Military:** There are several applications of ASR in military. Speech Recognition can be used to set radio frequencies, controlling weapon release parameters, and commanding an autopilot system. Specifically in the case of fighter jets and helicopters, the challenges associated with accurate speech recognition due to stressful environment and background acoustic noise.

**Technology:** Automation by virtual assistants is prevailing as a culture specifically on our mobile devices. Voice commands are used to access different functions and services on our smartphones and other devices such as
Apple's Siri, Google Assistant, Amazon's Alexa and Microsoft's Cortana. They are continuing to blend in our day-to-day life.

**Persons with Disabilities:** For people with hearing problems and having difficulty in writing, ASR can automatically create a text format of the give speech waveform such as discussions in conference, classroom or virtual lectures, and general public interactions.

## 3 Algorithms for Automated Speech Recognition

There have been several major innovations in the field of speech recognition. Various computation technique and algorithms have been proposed to convert speech into text with the fair accuracy of transcription. Recently, the availability of sufficient datasets and the advancement of Deep Learning (DL) have benefitted the field of

Automated Speech Recognition. DL impacted many Machines Learning (ML) fields that completely rely on domain knowledge. DL can replace decades of research of modeling with its DL models, which have higher accuracy and require less manual labor. There are some most commonly used methods for speech recognition listed below-

**Hidden Markov Models (HMMs):** Hidden Markov Models are statistical model based on the Markov process with unknown parameters, where the probability of the current state solely depends on current state and is independent of previous states. It is a type of Acoustic Model (AM) of speech recognition. In HMM, the states are not directly available but the variables which are influenced by the states are visible for inspection. Every single state has a probability distribution for the possible outcome tokens. HMM create stochastic models from the given known datasets and then compares the probabilities of the test or unknown datasets generated by each model.

HMM can be trained automatically and it is computationally feasible to use as it considers the audio signal as quasi-static for short durations, known as frames. It models these frames for speech recognition and then break the extracted feature vector from signal into a number of states and finds the probability of a signal to transit from one to another state. HMM are simple algorithm that can generate speech using a number of states for each model and correspondingly associated short-term spectra modelling. The parameters of model are the state transition probabilities and the means, variances and mixture weights, which characterize the state output distribution. HMM arrange the feature vector into a Markov chain which store probabilities of state transitions using the theory of statistics. It assigns labels to each unit in the sequence. These labels then create a mapping with the test input, allowing it to decide most probable label sequence.

**N-Gram:** N-Gram is the simplest Language Model (LM) that assigns probabilities to sentences and sequences of words. An N-Gram is a sequence of N-words: a 2-Gram or bigram is a two-word sequence like "come in" or "good day" similarly a 3-Gram for a three-word sequence like "please come in". It is a type of probabilistic LM used for predicting the next item of a sequence by looking back to (n-1) step.

An N-Gram model is built by analyzing the number of times, the word sequences appeared in a given dataset and based on this it estimates the probabilities. There are several limitations with N-Gram model. Smoothing, Interpolation and Back off are commonly used to improve the model.

**Artificial Neural Networks (ANNs):** Artificial Neural Networks are inspired by human brain's neural network system. A neural network system consists of interconnected layers consist of processing units, on the basis of training data different weights are given to different layers. ANNs try to mimic the human brain's behavior to perform speech recognition. Typically, ANN based model takes an acoustic input, process it through different layers and finally output the recognized text. The transcription is based the accuracy of weights decided through training datasets.

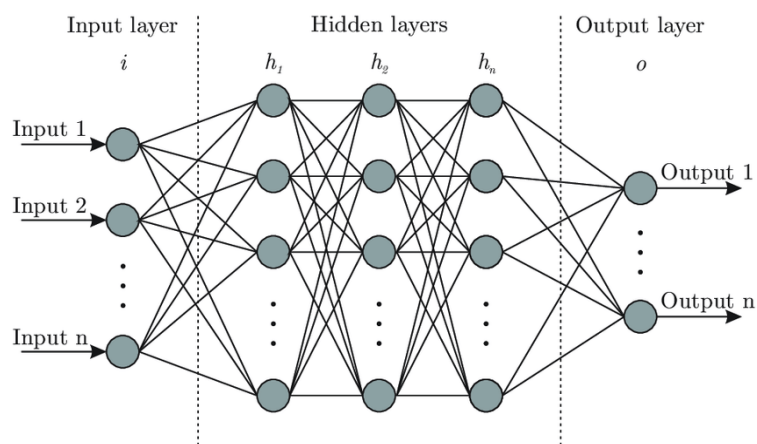

Fig: ANN architecture

Conventional Neural Network (CNN) of Multi-Layer Perceptron (MLP) type have been increasingly in use for modern day speech recognition. This type of neural network works very well for classifying the vowel sounds with stationary spectra, but poorly performs for phoneme discriminating of consonants, which are characterizes by variations of their short-term spectra. As feedforward multi-layer neural network are unable to cope up with the time varying information like time-varying spectra of speech sounds. This limitation is tackled by Recurrent Neural Networks (RNNs), which have ability of incorporating feedback structure by having connection between units of different layers and memorizing the time-varying information.

Classification and recognition of static patterns is the main advantage of ANNs, which is useful for speech recognition including noisy acoustic data from isolated speech units. However, only ANN-based systems do not perform well for the continuous speech units so ANNs are integrated with HMM in a hybrid way to get the optimized results.

## 4 Mozilla Deep Speech

Deep Speech is a state-of-the-art speech recognition system developed using Baidu's end-to-end ASR architecture. It is based on DL specifically on RNN, trained using multiple GPUs and large amount of speech data. As it learns directly from data, specialized techniques like speech adaptation or noise filtering are not required for it.

Traditional speech system mainly relies on heavily engineered processing stages and sophisticated pipelines composed of multiple algorithms including specialized features, acoustic models, and HMM. These pipelines need continuous modification for enhancing the performance thus great deal of effort tuning their features and models. Other than this, to improve the performance of speech recognition in the noisy environment, a specifically designed system is required for robustness. Deep Speech applies DL end-to-end using RNN. Large datasets provide the advantage for DL system to learn and improve overall performance. Deep Speech can learn robustness to noise or speaker variation automatically.

Deep Speech is a character level deep RNN, trained using end-to-end supervised learning. As features, it extracts s Mel Frequency Cepstral Coefficients from speech signals and then directly outputs the transcription. It does not require any external knowledge like a Grapheme to Phoneme (G2P) converter. The Deep Speech network consist of six layers: the speech features are fed into three fully connected layers (dense), then followed by a unidirectional RNN layer, then a fully connected layer (dense) and finally an output layer. The RNN layer uses Long Short-Term Memory (LSTM) cells, and the hidden fully connected layers use a ReLU activation function. The network outputs a matrix of character probabilities, meaning for each time step the system gives out a probability for each character in the alphabet, represents the likelihood of that particular character corresponding to the audio. Further, the Connectionist Temporal Classification (CTC) loss function is used to maximize the probability of the correct transcription.

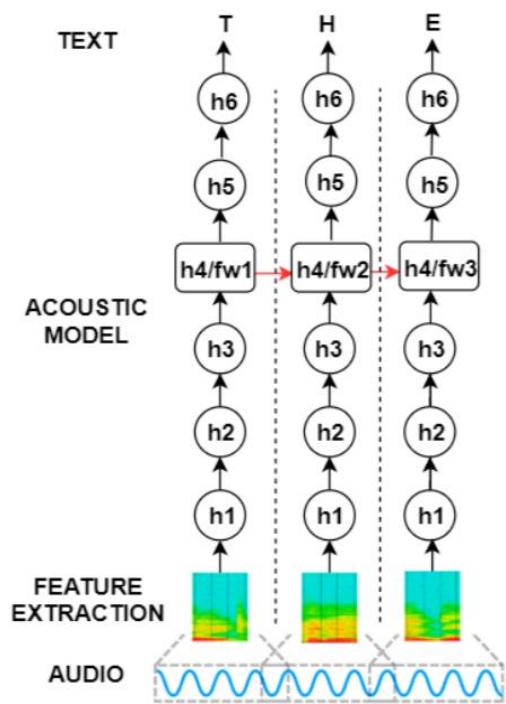

Fig: DeepSpeech architecture (source: Mozilla DeepSpeech blog)

## 5 Experiment Design and Training

This section will provide the details of the Deep Speech model training and fine tuning for Indian Accents English. Latest available Deep Speech model i.e., deepspeech-0.9.3 will be trained for six major English accents in India including both male and female speakers.

### 5.1 Datasets

Indic TTS dataset is utilized for the training purpose of Deep Speech. It is a multi-institutional project on developing text-to-speech (TTS) synthesis systems for Indian languages, improving quality of synthesis and TTS integrated various application. It is a consortium-based project funded by the Department of Electronics and Information Technology (Deity), Ministry of Communication and Information Technology, Government of India. Indic TTS English dataset consist of English language audios with their metadata file uttered by both male and female speakers from different regions and states of India. There are six English accents datasets namely Assamese, Bengali, Hindi, Kannada, Malayalam, and Tamil. Separate zip archives files 10000+ spoken sentences by both male and females in wav format and their corresponding transcription files. The total size of Indic TTS dataset for English accents is around 50 Gigabytes (GB).

### 5.2 Preprocessing

Deep Speech training and fine tuning requires data in a specific format so that they can be directly transfer into defined input pipelines of the algorithm. We need to process metadata file and generate a train, test, and dev split files for the Deep Speech training. Initially, an output file in csv format is generated, which is further divided into train, test, and dev split files. However, Indic TTS datasets consist of various irregularities which are not compatible with Deep Speech training. Preprocessing of Indic TTS data is required to make it ready for the training. There are several factors in the Indic TTS dataset generated csv file, which need to be taken care of like

**Sampling rate**: Sampling rate of audio files in dataset is 48 kHz, while required sampling rate for Deep Speech training is 16 kHz. We have down sampled each audio file to make it compatible for Deep Speech training.
**Special characters**: There are several special characters, punctuation marks, extra space and uppercase characters are there in the transcription files. We have optimized the transcription files by eradicating all these anomalies.

### 5.3 Augmentation and Hyperparameter Setup
Mozilla Deep Speech allows augmentation and hyperparameter setup for better generalization to the model training and fine-tuning. These are useful technique for the better generalization of ML models. It is used alter some training variables in the pre-processing pipelines. A corresponding flag is assigned to a particular augmentation, data will be trained by implying that particular augmentation technique. However, whether an augmentation will actually get applied to a sample training data is decided by chance on base of the augmentation's probability value. Most of the hyperparameters were set same as those in pre-configure in Mozilla Deep Speech but there are some changes made in batch size considering the amount of data for training and machine capacity. The best resulting hyperparameters are mentioned in the table below.

Besides augmentation and hyperparameter, Deep Speech also offer to create checkpoints during the training phase as due to some mishap, if the training stopped then it can be again started from that point by utilizing the checkpoints. Deep Speech gives you option to create a checkpoint directory for storing the training checkpoints and another load the Deep Speech checkpoint directory, where the checkpoints essential for training is located.

### 5.4 Training

For the initial training and testing with single language database, an open-source platform Google Colaboratory (Colab) is used. Colab provide the high computing performance by using GPU. A single gender of a particular language database took around one hour for training and validation with 30 epochs. A dedicated model trained on this particular database will be saved. This trained model performed better than the original Deep Speech model.

## 6 Analysis and Results

We have tested the trained model for the same accent type test dataset. We got Word Error Rate (WER) equal to 0.181523, Character Error Rate (CER) equal to 0.059941, and loss is 14.695212. The best performance we achieved, with WER is 0.00000 and CER is 0.00000 with loss 3.699683. The median performance was, when

WER equal to 0.157895, CER equal to 0.44444 with loss 11.959745. The worst performance was, when WER equal to 0.833333, CER equal to 0.142857 with loss 9.320448. Overall, we can infer from the data that the trained model tends to perform much better than the original Deep Speech model.

**7 Conclusion**

In this paper, a single end-to-end ASR system, trained on several English accents from all over the country is generated. This can be used in several application as discussed in earlier sections of this paper. It will act as a single platform for STT for the entire country.
Our results thus support the idea that Mozilla Deep Speech can be easily transferred to new English accents. No specific hardware is used to train and run this model. We are mentioning the configuration data for all experiments and setup in order to enable replicating all the results. As training script is made available, the model can be re-trained and optimised for the new English accents, which are not been used in this experiment. The usage and training can be done on normal desktop computer or laptop for limited amount of data.